\documentclass[runningheads,a4paper]{llncs}

\usepackage[utf8]{inputenc}
\usepackage[T2A]{fontenc}

\usepackage[russian,english]{babel}
\usepackage{amsmath}
\usepackage{graphicx}
\usepackage{booktabs}
\usepackage{cite}

\begin{document}

\title{Human and Machine Judgements for\\Russian Semantic Relatedness}
\titlerunning{Human and Machine Judgements for Russian Semantic Relatedness}

\author{
  Alexander Panchenko\inst{1} \and
  Dmitry Ustalov\inst{2} \and
  Nikolay Arefyev\inst{3} \and
  Denis Paperno\inst{4} \and
  Natalia Konstantinova\inst{5} \and
  Natalia Loukachevitch\inst{3} \and
  Chris Biemann\inst{1}
}

\authorrunning{Alexander Panchenko et al.}

\institute{
  TU Darmstadt, Darmstadt, Germany \\
  \email{\{panchenko,biem\}@lt.informatik.tu-darmstadt.de} \and
  Ural Federal University, Yekaterinburg, Russia \\
  \email{dmitry.ustalov@urfu.ru} \and
  Moscow State University, Moscow, Russia \\
  \email{louk\_nat@mail.ru} \and
  University of Trento, Rovereto, Italy \\
  \email{denis.paperno@unitn.it} \and
  University of Wolverhampton, Wolverhampton, UK \\
  \email{n.konstantinova@wlv.ac.uk}
}

\maketitle

\begin{abstract}\hyphenpenalty=10000\exhyphenpenalty=10000
Semantic relatedness of terms represents similarity of meaning by a numerical score. On the one hand, humans easily make judgements about semantic relatedness. On the other hand, this kind of information is useful in language processing systems. While semantic relatedness has been extensively studied for English using numerous language resources, such as associative norms, human judgements and datasets generated from lexical databases, no evaluation resources of this kind have been available for Russian to date. Our contribution addresses this problem. We present five language resources of different scale and purpose for Russian semantic relatedness, each being a list of triples $(\textit{word}_i, \textit{word}_j, \textit{similarity}_{ij}$). Four of them are designed for evaluation of systems for computing  semantic relatedness, complementing each other in terms of the semantic relation type they represent. These benchmarks were used to organise a shared task on Russian semantic relatedness, which attracted 19 teams. We use one of the best approaches identified in this competition to generate the fifth high-coverage resource, the first open distributional thesaurus of Russian. Multiple evaluations of this thesaurus, including a large-scale crowdsourcing study involving native speakers, indicate its high accuracy.
\end{abstract}

\section{Introduction}

Semantic relatedness numerically quantifies the degree of semantic alikeness of two lexical units, such as words and multiword expressions. The relatedness score is high for pairs of words in a semantic relation (e.g., synonyms, hyponyms, free associations) and low for semantically unrelated pairs. \textit{Semantic relatedness} and \textit{semantic similarity} have been extensively studied in psychology and computational linguistics, see~\cite{Budanitsky:06,Pedersen:07,Gabrilovich:07,Batet:11} inter alia. While both concepts are vaguely defined, similarity is a more restricted notion than relatedness, e.g. ``apple'' and ``tree'' would be related but not similar. Semantically similar word pairs are usually synonyms or hypernyms, while relatedness also can also refer to meronyms, co-hyponyms, associations and other types of relations. Semantic relatedness is an important building block of NLP techniques, such as text similarity~\cite{Bar:12,Tsatsaronis:10}, word sense disambiguation~\cite{Patwardhan:03}, query expansion~\cite{Hsu:06} and some others~\cite{Panchenko:13}.

While semantic relatedness was extensively studied in the context of the English language,  NLP researchers working with Russian language could not conduct such studies due to the lack of publicly available relatedness resources. The datasets presented in this paper are meant to fill this gap. Each of them is a collection of weighted word pairs in the format  $(w_i,w_j,s_{ij})$, e.g.~$(\textit{book}, \textit{proceedings}, 0.87)$. Here, the $w_i$ is the source word, $w_j$ is the destination word and $s_{ij} \in [0;1]$ is the semantic relatedness score (see \tablename~\ref{tab:relations}). 

More specifically, we present (1) four resources for evaluation and training of semantic relatedness systems varying in size and relation type and (2) the first open distributional thesaurus for the Russian language (see \tablename~\ref{tab:resources}). All datasets contain relations between single words.

The paper is organized as follows: Section~\ref{sec:related} describes approaches to evaluation of semantic relatedness in English.  Section~\ref{sec:human} presents three datasets where semantic relatedness of word was established manually. The HJ dataset, further described in Section~\ref{sub:hj}, is based on Human Judgements about semantic relatedness; the RuThes (RT) dataset is based on synonyms and hypernyms from a handcrafted thesaurus (see Section~\ref{sub:rt}); the Associative Experiment (AE) dataset, introduced in Section~\ref{sub:ae}, represents cognitive associations between words.  Section~\ref{sec:machine} describes datasets where semantic relatedness between words is established automatically: the Machine Judgements (MJ) dataset, presented in Section~\ref{sub:mj}, is based on a combination of submissions from a shared task on Russian semantic similarity; Section~\ref{sub:rdt} describes the construction and evaluation of the Russian Distributional Thesaurus (RDT).

\begin{table}[ht]
\scriptsize
\centering
\caption{Example of semantic relations from the datasets described in this paper: the five most and  least similar terms to the word ``книга'' (book) in the MJ dataset}
\label{tab:relations}
\begin{tabular}{lll}\toprule
\textbf{Source Word}, $w_j$ & \textbf{Destination Word}, $w_j$ & \textbf{Semantic Relatedness}, $s_{ij}$ \\ \midrule
книга (book) & книжка (book, little book)    & $0.719$ \\
книга (book) & книжечка (little book)        & $0.646$ \\
книга (book) & сборник (proceedings)         & $0.643$ \\
книга (book) & монография (monograph)        & $0.574$ \\
книга (book) & том (volume)                  & $0.554$ \\ \midrule
книга (book) & трест (trust as organization) & $0.151$ \\
книга (book) & одобрение (approval)          & $0.150$ \\
книга (book) & киль (keel)                   & $0.130$ \\
книга (book) & Марокко (Marocco)             & $0.124$ \\
книга (book) & Уругвай (Uruguay)             & $0.092$ \\ \bottomrule
\end{tabular}
\end{table}

\begin{table}[ht]
\centering
\caption{Language resources presented in this paper. The pipe ($|$) separates the sizes of two dataset versions: one with manual filtering of negative examples and the other version, marked by an asterix (*), where negative relations were generated automatically, i.e. without manual filtering}
\label{tab:resources}
\scriptsize
\begin{tabular}{l|lllll}\toprule
\textbf{Dataset}     & \textbf{HJ}  & \textbf{RT}            & \textbf{AE}           & \textbf{MJ}  & \textbf{RDT}    \\ \midrule
\# relations         & $398$        & $9\,548$ | $114\,066$* &  $3\,002$ | $86\,772$* & $12\,886$    & $193\,909\,130$ \\
\# source words, $w_i$      & $222$        & $1,008$ | $6\,832$*    & $340$ | $5\,796$*     & $1\,519$     & $931\,896$      \\
\# destination words, $w_j$ & $306$        & $7\,737$ | $71\,309$*  & $2\,498$ | $56\,686$* & $9\,044$     & $4\,456\,444$   \\
types of relations   & relatedness  & \parbox{15mm}{synonyms,\\%
                                      hypernyms}             & associations          & relatedness  & relatedness     \\
similarity score, $s_{ij}$     & from 0 to 1  & 0 or 1                 & 0 or 1                & from 0 to 1  & from 0 to 1     \\
part of speech        & nouns & nouns           & nouns          & nouns & any    \\ \bottomrule
\end{tabular}
\end{table}

\section{Related Work}\label{sec:related}

There are three main approaches to evaluating semantic relatedness: using human judgements about word pairs, using semantic relations from lexical-semantic resources, such as WordNet~\cite{Miller:95}, and using data from cognitive word association experiments. We built three evaluation datasets for Russian each based on one of these principles to enable a comprehensive comparison of relatedness models.

\subsection{Datasets Based on Human Judgements about Word Pairs}

Word pairs labeled manually on a categorical scale by human subjects is the basis of this group of benchmarks. High scores of subjects indicate that words are semantically related, low scores indicate that they are unrelated. The HJ dataset presented in Section~\ref{sub:hj} belongs to this group of evaluation datasets.

Research on relatedness starts from the pioneering work of Rubenstein and Goodenough~\cite{Rubenstein:65}, where they aggregated human judgments on the relatedness of $65$ noun pairs into the RG dataset. 51 human subjects rated the pairs on a scale from $0$ to $4$ according to their similarity. Later, Miller and Charles~\cite{Miller:91} replicated the experiment of Rubenstein and Goodenough, obtaining similar results on a subset of $30$ noun pairs. They used $10$ words from the high level (between $3$ and $4$), $10$ from the intermediate level (between $1$ and $3$), and $10$ from the low level ($0$ to $1$) of semantic relatedness, and then obtained similarity judgments from $38$ subjects, given the RG annotation guidelines, on those $30$ pairs. This dataset is known as the MC dataset. 

A larger set of $353$ word pairs was put forward by Filkenstein et al.~\cite{Finkelstein:01} as the WordSim353 dataset. The dataset contains $353$ word pairs, each associated with $13$ or $16$ human judgements. In this case, the subjects were asked to rate word pairs for relatedness, although many of the pairs also exemplify semantic similarity. That is why, Agirre et al.~\cite{Agirre:09} subdivided the WordSim353 dataset into two subsets: the WordSim353 similarity  set and the WordSim353 relatedness set. The former set consists of word pairs classified as synonyms, antonyms, identical, or hyponym-hypernym and unrelated pairs. The relatedness set contains word pairs connected with other relations and unrelated pairs. The similarity set contains $204$ pairs and the relatedness set includes $252$ pairs.

The three abovementioned  datasets were created for English. There have been several attempts to translate those datasets into other languages. Gurevych translated the RG and MC datasets into German~\cite{Gurevych:05}; Hassan and Mihalcea translated them into Spanish, Arabic and Romanian~\cite{Hassan:09}; Postma and Vossen~\cite{Postma:14} translated the datasets into Dutch; Jin and Wu~\cite{Jin:12} presented a shared task for Chinese semantic similarity, where the authors translated the WordSim353 dataset. Yang and Powers~\cite{Yang:06} proposed a dataset specifically for measuring verb similarity, which was later translated into German by Meyer and Gurevych~\cite{Meyer:12}.

Hassan and Mihalcea~\cite{Hassan:09} and Postma and Vossen~\cite{Postma:14} used three stages to  translation pairs: (1) disambiguation of the English word forms; (2) translation for each word; (3) ensuring that translations are in the same class of relative frequency as the English source word.

More recently, SimLex-999 was released by Hill et al.~\cite{Hill:15}, focusing specifically on similarity and not relatedness. While most datasets are only available in English, SimLex-999 became a notable exception and has been translated into German, Russian and Italian. The Russian version of SimLex-999 is similar to the HJ dataset presented in our paper. In fact, these Russian datasets were created in parallel almost at the same time\footnote{The HJ dataset was first released in November 2014 and first published in June 2015, while the SimLex-999 was first published December 2015.}. SimLex-999 contains 999 word pairs, which is considerably larger than the classical MC, RG and WordSim353 datasets. 

The creators of the MEN dataset~\cite{Bruni:14} went even further, annotating via crowdsourcing 3\,000 word pairs sampled from the ukWaC corpus~\cite{Ferraresi:08}. However, this dataset is also  available only for English. A comprehensive list of datasets for evaluation of English semantic relatedness, featuring 12 collections, was gathered by Faruqui and Dyer~\cite{Faruqui:14}. This set of benchmarks was used to build a web application for evaluation and visualization of word vectors.\footnote{\url{http://wordvectors.org/suite.php}}


\subsection{Datasets Based on Lexical-Semantic Resources}

Another group of evaluation datasets evaluates semantic relatedness scores with respect to relations described in lexical-semantic resources such as WordNet. The RT dataset presented in Section~\ref{sub:rt} belongs to this group of evaluation datasets.

Baroni and Lenci~\cite{Baroni:11} stressed that semantically related words differ in the type of relation between them, so they generated the BLESS dataset containing tuples of the form $(w_j, w_j, \textit{type})$. Types of relations included co-hyponyms, hypernyms, meronyms, attributes (relation between a noun and an adjective expressing its attribute), event (relation between a noun and a verb referring to actions or events). BLESS also contains, for each target word, a number of random words that were checked to be semantically unrelated to the this word. BLESS includes $200$ English concrete single-word nouns having reasonably high frequency that are not very polysemous. The destination words of the non-random relations are English nouns, verbs and adjectives selected and validated manually using several sources including WordNet, and collocations from the Wikipedia and the ukWaC corpora. 

Van de Cruys~\cite{vanDeCruys:10} used Dutch WordNet to evaluate distributional similarity measures. His approach uses the structure of the lexical resource, whereby distributional similarity is compared to shortest-path-based distance. Biemann and Riedl~\cite{Biemann:13} follow a similar approach based on the English WordNet to assess quality of their distributional semantics framework. 

Finally, Sahlgren~\cite{Sahlgren:06} evaluated distributional lexical similarity measures comparing them to manually-crafted thesauri, but also associative norms, such as those described in the following section. 

\subsection{Datasets Based on Human Word Association Experiments}

The third strain of research evaluates the ability of current automated systems to simulate the results of human word association experiments. Evaluation tasks based on associative relations originally captured attention of psychologists, such as Griffiths and Steyvers~\cite{Griffiths:03}. One such task was organized in the framework of the Cogalex workshop~\cite{Rapp:14}. The participants received lists of five  words (e.g.  ``circus'', ``funny'', ``nose'', ``fool'', and ``Coco'') and were supposed to select the word most closely associated to all of them. In this specific case, the word ``clown'' is the expected response. $2\,000$ sets of five input words, together with the expected target words (associative responses) were provided as a training set to participants. The test dataset contained another $2\,000$ sets of five input words. The training and the test datasets were both derived from the Edinburgh Associative Thesaurus (EAT)~\cite{Kiss:73}. For each stimulus word, only the top five associations, i.e. the associations produced by the largest number of respondents, were retained, and all other associations were discarded. The AE dataset presented in Section~\ref{sub:ae} belongs to this group of evaluation datasets.

\section{Human Judgements about Semantic Relatedness}\label{sec:human}

In this section, we describe  three datasets designed for evaluation of Russian semantic relatedness measures. The datasets were tested in the framework of the  shared task on RUssian Semantic Similarity Evaluation (RUSSE)~\cite{Panchenko:15}.\footnote{\url{http://russe.nlpub.ru}} Each participant had to calculate similarities between a collection of word pairs. Then, each submission was assessed using the three benchmarks presented below, each being a subset of the input word pairs.

\subsection{HJ: Human Judgements of Word Pairs}\label{sub:hj}

\subsubsection{Description of the Dataset.} The HJ dataset is a union of three widely used benchmarks for English: RG, MC and WordSim353, see~\cite{Resnik:95,Lin:98,Patwardhan:06,Patwardhan:06,Zesch:08,Agirre:09} inter alia. The dataset contains 398 word pairs translated to Russian and re-annotated by native speakers. 
In addition to the complete dataset, we also provide separate  parts that correspond to MC, RG and WordSim353.

To collect human judgements,  an in-house crowdsourcing system was used. We set up a special section on the RUSSE website and asked volunteers on Facebook and Twitter
to participate in the experiment. Each annotator received an assignment consisting of 15 word pairs randomly selected from the 398 pairs, and has been asked to assess the relatedness of each pair on the following scale: 0 -- not related at all, 1 -- weak relatedness, 2 -- moderate relatedness, and 3 -- high relatedness. We provided
annotators with simple instructions explaining the procedure and goals of the study.\footnote{Annotation guidelines for the HJ dataset: \url{http://russe.nlpub.ru/task/annotate.txt}} 

A pair of words was added to the annotation task with the probability inversely proportional to the number of current
annotations. We obtained a total of 4\,200 answers, i.e. 280 submissions of 15 judgements. Ordinal Krippendorff's alpha of $0.49$ indicates a moderate agreement of annotators. The scores included in the HJ dataset are average human ratings scaled to the $[0,1]$ range.



\subsubsection{Using the Dataset.} To evaluate a relatedness measure using this dataset one should (1) calculate relatedness scores for each pair in the dataset; (2) calculate Spearman's rank correlation coefficient $\rho$ between the vector of human judgments and the scores of the  system (see \tablename~\ref{tab:evaluation} for an example). 

\subsection{RT: Synonyms and Hypernyms}\label{sub:rt}

\subsubsection{Description of the Dataset.} This dataset follows the structure of the BLESS dataset \cite{Baroni:11}. Each target word has the same number of related and unrelated source words. The dataset contains $9\,548$ relations for $1\,008$ nouns (see \tablename~\ref{tab:resources}). Half of these relations are synonyms and hypernyms from the RuThes-lite thesaurus \cite{Loukachevitch:14} and half of them are unrelated words. To generate negative pairs we used the automatic procedure described in Panchenko et al. \cite{Panchenko:15}. We filtered out false negative relations for $1\,008$ source words with the help of human annotators. Each negative relation in this subset was annotated by at least two annotators: Masters' students of an NLP course, native speakers of Russian. 

As the result, we provide a dataset featuring $9\,548$ relations of 1\,008 source words, where  each source word has the same number of negative random relations and positive (synonymous or hypernymous) relations. In addition, we provide a larger dataset of $114\,066$ relations for $6\,832$ source words, where negative relations have not been verified manually.

\subsubsection{Using the Dataset.} To evaluate a similarity measure using this dataset one should (1) calculate relatedness scores for each pair in the dataset; (2) first sort pairs by the score; and then (3) calculate the average precision metric: 
$$
AveP = \frac{\sum_r P(r)}{R},
$$
where $r$ is the rank of each non-random pair, $R$ is the total number of non-random pairs, and $P(r)$ is the precision of the top-$r$ pairs. See \tablename~\ref{tab:evaluation} and \cite{Panchenko:15} for examples. Besides, the dataset can be used to train classification models for predicting hypernyms and synonyms using the binary $s_{ij}$ scores. 

\subsection{AE: Cognitive Associations}\label{sub:ae}

\subsubsection{Description of the Dataset.} The structure of this dataset is the same as the structure of the RT dataset: each source word has the same number of related and unrelated target words. The difference is that, related word pairs of this dataset were sampled from a Russian web-based associative experiment.\footnote{The associations were sampled from the \url{sociation.org} database in July 2014. } In the experiment, users were asked to provide a reaction to an input stimulus source word, e.g.: man $\rightarrow$ woman, time $\rightarrow$ money, and so on. The strength of association in this experiment is quantified by the number of respondents providing the same stimulus-reaction pair. Associative thesauri typically contain a mix of synonyms, hyponyms, meronyms and other types, making relations asymmetric. To build this dataset, we  selected target words with the highest association with the stimulus in \url{Sociation.org} data. Like with the other datasets, we used only single-word nouns. Similarly to the RT dataset, we automatically generated negative word pairs and filtered out false negatives with help of annotators.

As the result, we provide a dataset featuring $3\,002$ relations of $340$ source words (see \tablename~\ref{tab:resources}), where  each source word has the same number of negative random relations and positive associative relations. In addition, we provide the larger dataset of $86\,772$ relations for $5\,796$ source words, where negative relations were not verified manually.

\subsubsection{Using the Dataset.}  Evaluation procedure using this dataset is the same as for the RT dataset: one should calculate the average precision $AveP$. Besides, the dataset can be used to train classification models for predicting associative relations using the binary $s_{ij}$ scores.

\section{Machine Judgements about Semantic Relatedness}\label{sec:machine}

\subsection{MJ: Machine Judgements of Word Pairs }\label{sub:mj}

\subsubsection{Description of the Dataset.}

This dataset contains $12\,886$ word pairs of $1\,519$ source words coming from HJ, RT, and AE datasets. Only $398$ word pairs from the HJ dataset have continuous scores, while the other pairs which come from the RT and the AE datasets have binary relatedness scores. However, for training and evaluation purposes it is desirable to have continuous relatedness scores as they distinguish between the shades of relatedness. Yet, manual annotation of a big number of pairs is problematic: the largest dataset of this kind available to date, the MEN, contains $3\,000$ word pairs. Thus, unique feature of the MJ dataset is that it is at the same time large-scale, like BLESS, and has accurate continuous scores, like WordSim-353. 

To estimate continuous relatedness scores with high confidence without any human judgements, we used $105$ submissions of the shared task on Russian semantic similarity (RUSSE). We assumed that the top-scored systems can be used to bootstrap relatedness scores. Each run of the shared task consisted of $12\,886$ word pairs along with their relatedness scores. We used the following procedure to average these scores and construct the dataset:

\begin{enumerate}
  \item Select one best submission for each of the $19$ participating teams for HJ, RT and AE datasets (total number of submissions is 105).
  \item Rank the $n=19$ best submissions according to their results in HJ, RT and AE: $r_k = n + 1 - k$, where $k$ is the place in the respective track. The best system obtains the rank $r_1=19$; the worst one has the rank $r_{19}=1$.
  \item Combine scores of these 19 best submissions as follows: $s'_{ij}=\frac{1}{n} \sum^{n}_{k=1} \alpha_k s^{k}_{ij}$, where $s^{k}_{ij}$ is the similarity between words $(w_i, w_j)$ of the $k$-th submission; $\alpha_k$ is the weight of the $k$-th submission. We considered three combination strategies each discounting differently teams with low ranks in the final evaluation. Thus the best teams impact more the combined score. In the first strategy, the $\alpha_k$ weight is the rank $r_k$. In the second strategy, the $\alpha_k$ equals exponent of this rank: $\exp(r_k)$. Finally, in the third strategy, the weight equals to the  square root of rank: $\sqrt{r_k}$. We tried to  use  $AveP$ and $\rho$ as weights, but this did not lead to better fit.
  \item Union pairs $(w_i, w_j, s'_{ij})$ of HJ, RT and AE datasets into the MJ dataset. \tablename~\ref{tab:relations} presents example of the relatedness scores obtained using this procedure. 
\end{enumerate}

\subsubsection{Evaluation of the Dataset.}

Combination of the submissions using any of the three methods yields relatedness scores that outperforms all single submissions of the shared task (see \tablename~\ref{tab:ranks}). Note that ranks of the systems were obtained using the HJ, RT and AE datasets. Thus we can only claim that MJ provides continuous relatedness scores that fit well to the binary scores. Among the three weightings, using inverse ranks  provides the top scores on the HJ and the AE datasets and the second best scores on the RT dataset. Thus, we selected this strategy to generate the released dataset. 

\subsubsection{Using the Dataset.} To evaluate a relatedness measure using the MJ dataset, one should (1) calculate relatedness scores for each pair in the dataset; (2) calculate Spearman's rank correlation $\rho$ between the vector of machine judgments and the scores of the evaluated system. Besides, the dataset can be used to train regression models for predicting semantic relatedness using the continuous $s_{ij}$ scores.

\begin{table}[ht]
\centering
\caption{Performance of three combinations of submissions of the RUSSE shared task compared to the best scores for the HJ/RT/AE datasets across all submissions}
\label{tab:ranks}
\begin{tabular}{l|ccc}\toprule
                          & \textbf{HJ, $\rho$} & \textbf{RT, $AveP$} & \textbf{AE, $AveP$} \\ \midrule
The best RUSSE submissions for resp. datasets & $0.762$             & $0.959$             & $0.985$             \\
\bf MJ: $\alpha_k$ is the rank $r_k$                  & $\mathbf{0.790}$    & $0.990$             & $\mathbf{0.992}$    \\
MJ:$\alpha_k$ is the exponent of rank $\exp(r_k)$      & $0.772$             & $\mathbf{0.996}$    & $0.991$             \\
MJ: $\alpha_k$ is the sqrt of rank $\sqrt{r_k}$          & $0.778$             & $0.983$             & $0.989$ \\ \bottomrule
\end{tabular}
\end{table}

\subsection{RDT: Russian Distributional Thesaurus}\label{sub:rdt}

While four resources presented above are accurate and represent different types of semantic relations, their coverage (222 -- 1\,519 source words) makes them best suited for evaluation and training purposes. In this section, we present a large-scale resource in the same $(w_i, w_j, s_{ij})$ format, the first open Russian distributional thesaurus. This resource, thanks to its coverage of $932\,896$ target words can be directly used in NLP systems.

\subsubsection{Description of the Dataset.}

In order to build the distributional thesaurus, we used the Skip-gram model \cite{Mikolov:13} trained on a 12.9 billion word collection of Russian texts extracted from the digital library \url{lib.rus.ec}. According to the results of the shared task on Russian semantic relatedness \cite{Panchenko:15,Arefyev:15}, this approach scored in the top $5$ among $105$ submissions, obtaining different ranks depending on the evaluation dataset. At the same time, this method is completely unsupervised and language independent as we do not use any preprocessing except tokenization, in contrast to other top-ranked methods e.g.~\cite{Lopukhin:15} who used extra linguistic resources, such as dictionaries.

Following our prior experiments \cite{Arefyev:15}, we  selected the following parameters of the model: minimal word frequency -- $5$, number of dimensions in a word vector -- $500$, three or five iterations of the learning algorithm over the input corpus, context window size of $1$, $2$, $3$, $5$, $7$ and $10$ words. We calculated $250$ nearest neighbours using the cosine similarity between word vectors for the 1.1 million of the most frequent tokens. Next we filtered all tokens with non-Cyrillic symbols which provided us a resource featuring $932\,896$ source words. In addition to the raw tokens we provide a lemmatized version based on the PyMorphy2 morphological analyzer~\cite{Korobov:15}. We performed no part of speech filtering as it can be trivially performed if needed.  

\begin{figure}
\centering
\includegraphics[width=1.0\textwidth]{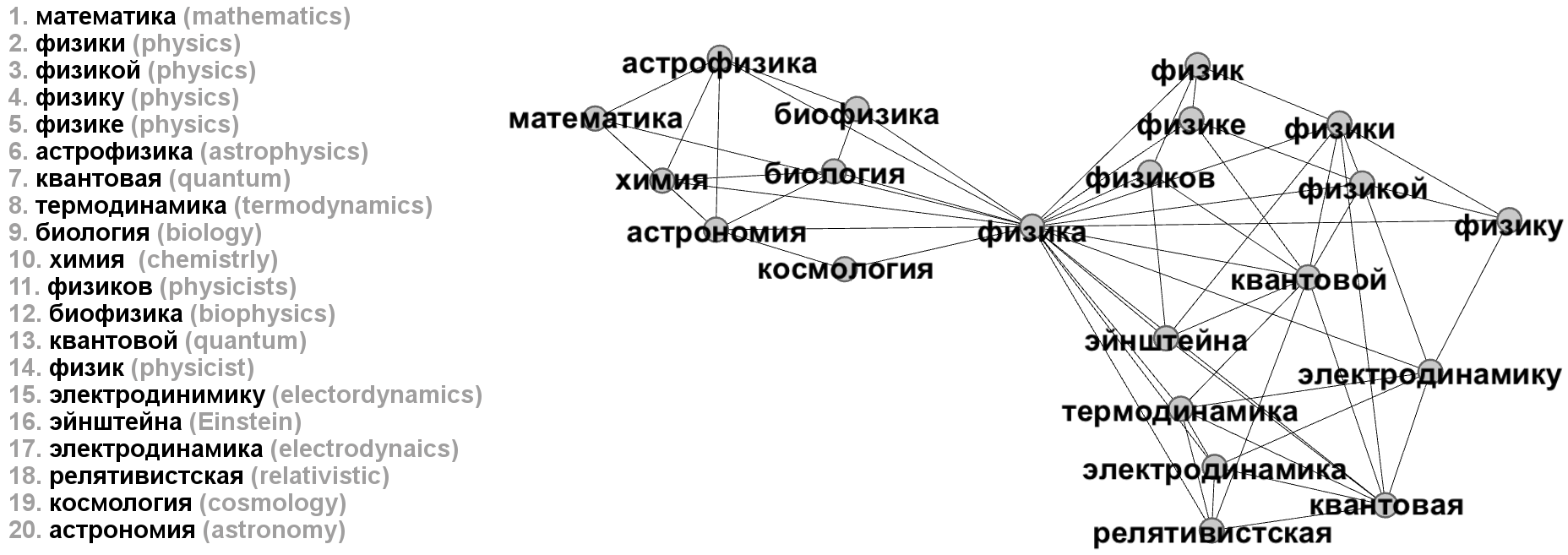}
\caption{Visualization of the 20 most semantically related words to the word ``физика'' (physics) in the Russian Distributional Thesaurus in the form of a list (on the left) and an ego-network (on the right)}
\label{fig:physics}
\end{figure}

\figurename~\ref{fig:physics} visualizes top 20  nearest neighbours of the word ``физика'' (physics) from the RDT. One can observe three groups of related words: morphological variants (e.g. ``физике'', ``физику''), physical terms, e.g. ``квантовая'' (quantum) and ``термодинамика'' (thermodynamics) and names of other scientific disciplines, e.g. ``математика'' (mathematics), ``химия'' (chemistry). Note that the thesaurus contains both raw tokens as displayed in \figurename~\ref{fig:physics} and lemmatized neighbours. 

An important added value of our work is engineering. While our approach is straightforward, training a large-scale Skip-gram model on a 12.9 billion tokens corpus with three iterations over a corpus takes up to five days on a \texttt{r3.8xlarge} Amazon EC2 instance featuring $32$ CPU cores and $244$ GB of RAM. Furthermore, computation of the neighbours takes up to a week for only one model using the large 500 dimensional vectors, not to mention the time needed to test different configurations of the model. Besides, to use the word embeddings directly, one needs to load more than seven millions of the 500 dimensional vectors, which is only possible on a similar instance to \texttt{r3.8xlarge}. On the other hand, the resulting RDT resource is a CSV file that can be easily indexed in an RDBMS system or an succinct in-memory data structure and subsequently efficiently used in most NLP systems. However, we also provide the original word vectors for non-standard use-cases. 

\subsubsection{Evaluation.} We evaluated the quality of the distributional thesaurus using the  HJ, RT and AE datasets presented above. Furthermore, we estimated precision of extracted relations for $100$ words randomly sampled from the vocabulary of the HJ dataset. For each word we extracted the top $20$ similar words according to each model under evaluation resulting in $4\,127$ unique word pairs. Each pair was annotated by three distinct annotators with a binary choice as opposed to a graded judgement, i.e.\ an annotator was supposed to indicate if a given word pair is plausibly related or not.\footnote{Annotation guidelines are available at \url{http://crowd.russe.nlpub.ru}.} In this experiment, we used an open source crowdsourcing engine \cite{Ustalov:15:ispras}.\footnote{\url{http://mtsar.nlpub.org}}
Judgements were aggregated using a majority vote. In total, $395$ Russian-speaking volunteers participated in our crowdsourcing experiment with the substantial inter-rater agreement of $0.47$ in terms of Krippendorff's alpha. The dataset obtained as a result of this crowdsourcing is publicly available (see download link below). 

\subsubsection{Discussion of the Results.} 
\begin{figure}[ht]
\centering
\includegraphics[width=.9\textwidth]{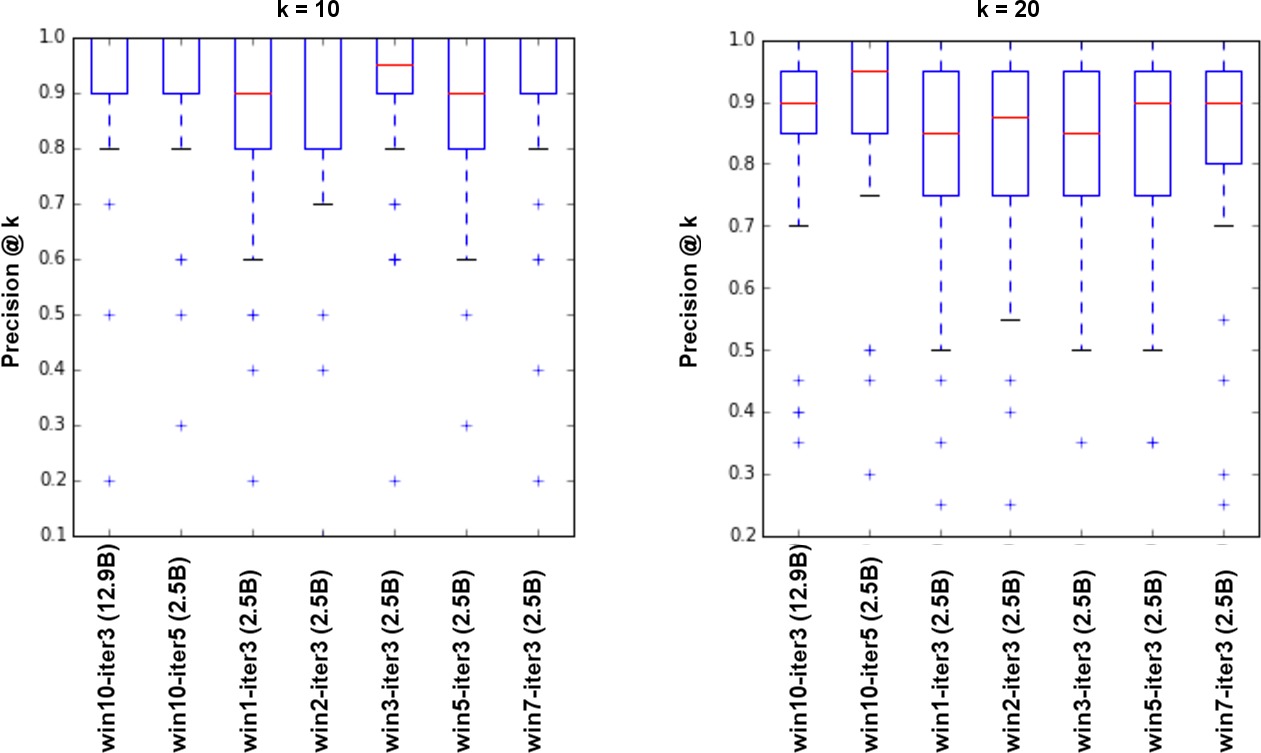}
\caption{Precision at $k \in \{10,20\}$  top similar words of the RDT based on the Skip-gram model with $500$ dimensions evaluated using crowdsourcing. The plot shows dependence of the performance of size of the context window (window size $1-10$) and size of the training corpus ($2.5$ and $12.9$ billions of tokens) and number of iterations during training ($3$ or $5$)}
\label{fig:box}
\end{figure}

Evaluation of different configurations of the distributional thesaurus are presented in \tablename~\ref{tab:evaluation} and \figurename~\ref{fig:box}. The model trained on the full 12.9 billion tokens corpus with context window size $10$ outperforms other models according to HJ, RT, AE and precision at 20 metrics. We used this model to generate the thesaurus presented in \tablename~\ref{tab:resources}. However, the model trained on the 2.5 billion tokens sample of the full lib.rus.ec corpus (20\% of the full corpus) yields very similar results in terms of precision. Yet, this model show slightly lower results according to other benchmarks. Models based on other context window sizes yield lower results as compared to these trained using the context window size 10 (see \figurename~\ref{fig:box}).


\begin{table}[ht]
\centering
\caption{Evaluation of different configurations of the Russian Distributional Thesaurus (RDT). The upper part of the table reports performance based on correlations with human judgements (HJ), semantic relations from a thesaurus (RT), cognitive associations (AE) and manual annotation of top 20 similar words assessed with precision at $k$ ($P@k$). The lower part of the table reports result of the top 4 alternative approaches from the RUSSE shared task}
\label{tab:evaluation}
\footnotesize
\begin{tabular}{lc|ccc|cccc}\toprule
\textbf{Model} & \textbf{\#tok.} & \textbf{HJ, $\rho$} & \textbf{RT, $AvgP$} & \textbf{AE, $AveP$} & $\mathbf{P@1}$ & $\mathbf{P@5}$ & $\mathbf{P@10}$ & $\mathbf{P@20}$ \\ \midrule
\texttt{win10-iter3} & $12.9$B & $\mathbf{0.700}$ & $\mathbf{0.918}$ & $\mathbf{0.975}$ & $0.971$          & $\mathbf{0.971}$ & $0.944$          & $\mathbf{0.912}$ \\
\texttt{win10-iter5} & $2.5$B  & $0.675$          & $0.885$          & $0.970$          & $\mathbf{1.000}$ & $\mathbf{0.971}$ & $\mathbf{0.947}$ & $0.910$          \\
\texttt{win5-iter3}  & $2.5$B  & $0.678$          & $0.886$          & $0.966$          & $\mathbf{1.000}$ & $0.953$          & $0.935$          & $0.881$          \\
\texttt{win3-iter3}  & $2.5$B  & $0.680$          & $0.887$          & $0.959$          & $0.971$          & $0.953$          & $0.935$          & $0.884$          \\ \midrule
\texttt{5-rt-3}~\cite{Lopukhin:15}  & --  & \bf 0.763          & \bf 0.923          & \bf 0.975          & --          & --          & --          & --          \\
\texttt{9-ae-9}~\cite{Panchenko:15}  & --  & $0.719$          & $0.884$          & $0.952$          & --          & --          & --          & --          \\
\texttt{9-ae-6}~\cite{Panchenko:15}  & --  & $0.704$          & $0.863$          & $0.965$          & --          & --          & --          & --          \\
\texttt{17-rt-1}~\cite{Panchenko:15}  & --  & $0.703$          & $0.815$          & $0.950$          & --          & --          & --          & --          \\
\bottomrule
\end{tabular}
\end{table}

\section{Conclusion}

In this paper, we presented five new language resources for the Russian language, which can be used for training and evaluating  semantic relatedness measures, and to create NLP applications requiring semantic relatedness. These resources were used to perform a large-scale evaluation of $105$ submissions in a shared task on Russian semantic relatedness. One of the best systems identified in this evaluation campaign was used to generate the first open Russian distributional thesaurus. Manual evaluation of this thesaurus, based on a large-scale crowdsourcing with native speakers, showed a precision of $0.94$ on the top $10$ similar words. All introduced resources are freely available for download.\footnote{\url{http://russe.nlpub.ru/downloads}} Finally, the methodology for bootstrapping datasets for semantic relatedness presented in this paper can help to construct similar resources in other languages.  

\subsubsection{Acknowledgements.} We would like to acknowledge several funding organisations that partially supported this research. Dmitry Ustalov was supported by the Russian Foundation for Basic Research (RFBR) according to the research project no.~16-37-00354~мол\_а. Denis Paperno was supported by the European Research Council (ERC) 2011 Starting Independent Research Grant no. 283554 (COMPOSES). Natalia Loukachevitch was supported by Russian Foundation for Humanities (RFH), grant no.~15-04-12017. Alexander Panchenko was supported by the Deutsche For\-schungs\-gemeinschaft (DFG) under the project ``Joining Ontologies and Semantics Induced from Text (JOIN-T)''.

\bibliographystyle{splncs}
\bibliography{rsr.aist2016}

\end{document}